\documentclass[10pt,letterpaper]{article}

\usepackage[top=0.85in,left=2.75in,footskip=0.75in]{geometry}

\usepackage{amsmath,amssymb}
\usepackage{changepage}
\usepackage[utf8x]{inputenc}

\usepackage{textcomp,marvosym}
\usepackage{cite}
\usepackage{nameref,hyperref}
\usepackage[right]{lineno}
\usepackage{microtype}
\DisableLigatures[f]{encoding = *, family = * }
\usepackage[table]{xcolor}
\usepackage{array}
\usepackage{xspace}
\usepackage{booktabs}
\usepackage{todonotes}

\newcolumntype{+}{!{\vrule width 2pt}}

\newlength\savedwidth

\usepackage{setspace} 
\raggedright
\usepackage[aboveskip=1pt,labelfont=bf,labelsep=period,justification=raggedright,singlelinecheck=off]{caption}

\newcommand{\SD}{\hphantom{$-$}}
\newcommand{\SN}{\hphantom{0}}

\makeatletter
\renewcommand{\@biblabel}[1]{\quad#1.}
\makeatother

\date{}

\usepackage{lastpage,fancyhdr,graphicx}
\usepackage{epstopdf}
\fancyheadoffset[L]{2.25in}
\fancyfootoffset[L]{2.25in}

\begin{document}
\vspace*{0.2in}

\begin{flushleft}
{\Large
\textbf\newline{Evaluation of Automatic Video Captioning Using Direct Assessment} 
}
\newline
\\
Yvette Graham$^\dagger$ \qquad
George Awad$^\ddagger$ \qquad
Alan F. Smeaton$^\ast$
\\
\bigskip
$^\dagger$ ADAPT Research Centre, Dublin City University, Glasnevin, Dublin 9, Ireland
\\
$^\ddagger$ National Institute of Standards and Technology, Gaithersburg, MD, USA; Dakota Consulting, Inc.
\\
$^\ast$ Insight Centre for Data Analytics, Dublin City University, Glasnevin, Dublin 9, Ireland
\\
\bigskip

%
%

Contact: graham.yvette@gmail.com

\end{flushleft}
\section*{Abstract}

We present Direct Assessment, a method for manually assessing the quality of automatically-generated  captions for video.  Evaluating the accuracy of video captions is particularly difficult because for any given video clip there is no definitive ground truth or correct answer against which to measure.  Automatic metrics for comparing automatic video captions against a manual caption such as BLEU and METEOR, drawn from techniques used in evaluating machine translation, were used in the TRECVid video captioning task in 2016 but these are shown to have weaknesses.  The work presented here brings human assessment into the evaluation by crowdsourcing how well a caption describes a video.  We automatically degrade the quality of some sample captions which are assessed manually and from this we are able to rate the quality of the human assessors, a factor we take into account in the evaluation. Using data from the TRECVid video-to-text task in 2016, we show how our direct assessment method is replicable and robust and should scale to where there many caption-generation techniques to be evaluated.



\section*{Introduction}


Describing image content, either still image or video, is difficult because images are rich in content and because they can be interpreted in so many ways.  Recent years have seen huge growth in the ways in which we can access imagery online.  Services like Netflix and Amazon Prime now bring moving images into our homes and onto our devices and social media services like Facebook and Twitter have created new ways in which we find this content. 

Current approaches to describing internet images and video is by {\em tagging}, which is identifying a set of objects or activities present in an image or video.  Tagging is not an exhaustive process in that it does not provide complete description of content.  While images and videos can be tagged manually, as a result of developments in machine learning and in particular deep learning, we can now do this automatically with good accuracy, with a wide spread of objects and activities, and with reasonable computation overhead \cite{GUO201627}.

There is a large scientific activity in the multimedia community to evaluate the effectiveness of automatic image and video descriptions  with benchmarking activities and data challenges. These include the semantic indexing task for video in TRECVid \cite{awad2016trecvida, smeaton2006evaluation},  the detection and classification of hundreds of object categories on millions of images in
ImageNet \cite{russakovsky2015imagenet} and ImageCLEF addressing the evaluation of image retrieval \cite{muller2010experimental}. Benchmarking activities like these are now mainstream within the multimedia and multimedia retrieval communities.

When it comes to describing video content, tagging becomes less attractive than it is for images.  This is because objects in a video can appear and disappear or be occluded during the running of a video. Thus tagging, if it was to be used, would need to be temporal and perhaps localised or boxed within video frames to show the objects that are tagged.  Tracking of tagged objects throughout a video shot might then be done based on bags of tags, one bag for each frame or object appearance.  More problematic is detecting and then describing the {\em activities} that happen in a video, the actions and movements, where simple tags are not rich or powerful enough to capture those activities or the relationships. For example temporal relationships like somebody did something {\em and then} something else happened, or  spatial relationships like somebody did something while {\em behind them} something else happened.  Such relationships are just  not suited to being described by simple tags.  

The alternative representation for videos is captions, descriptive narrative text which explains or describes what is happening in a video. The problem here is that by its nature, videos can be so rich in terms of their content and there can be so much happening in a short clip, that unlike image tagging where there is a groundtruth against which to measure, there is no single correct caption for a video.

In this paper we present a way to evaluate the accuracy and effectiveness of automatically-generated captions for short video clips.  Our approach uses human assessments as part of the evaluation process and our work is carried out in the context of the 2016 TRECVid Video-To-Text (VTT) pilot track.  The paper is organised as follows. In the next section we motivate the work by showing the need for having a human-in-the-loop for evaluations where there is no gold standard or groundtruth against which to measure automatically. We then describe how in previous work we have involved human assessment in evaluations of the accuracy of machine translation.  Section 4 gives an overview of the TRECVid VTT task, the data used, runs submitted by participants and metrics used while in Section 5 we describe our contribution, a method to incorporate human assessments into video captioning. Finally, in section 6 we describe our plans for using this human-in-the-loop for future evaluations.


\section*{Evaluations Using Humans-in-the-loop}

Given that we know it is difficult to evaluate generating video captions, either manual or automatically generated,  when there is no absolute correct answer, and thus nothing to measure new systems against this creates problems in terms of scientific validation.   Scientific research demands evaluations which are reliable and reproducible and thus independently verifiable and we want to achieve this with automatic captioning so we can see which techniques and approaches work best. While this may appear to be an intractable situation, there are other areas of scientific research where even without groundtruth, evaluations can be carried out, and one of those is in social media research where there is a huge surge of interest because of the importance of social media in our lives and the data is now easily available. 

Research into social media is usually to determine some form of behavioral patterns, such as sentiment towards issues or products, or the distribution of demographics interested in some topic, or tracking the spread, or truthfulness, associated with news.  Validating such patterns can involve tracking individual users and surveying or interacting with them but social media users are scattered and generally unavailable for one-on-one interaction or for providing feedback to researchers To validate the patterns they are trying to determine or to prove.  

This evaluation dilemma --- how to prove research findings in a credible and reproducible way when there is no ground truth --- is addressed in \cite{Zafarani:2015:EWG:2783419.2666680} which describes a number of techniques used in social media research. These include A/B testing such as the 2014 A/B test that Facebook carried out on almost 700,000 of its users investigating the transference of emotional state \cite{kramer2014experimental}, as well as spatiotemporal evaluation where predictions of future events or postings made using machine learning can be assessed by partitioning data into training and testing data, or tracking down the trigger for some event or phenomenon like a news activity or increased network traffic in some area by examining and exploring the counterfactual, namely what would have happened if the trigger event had not happened.

In the case of work reported in this paper there isn't a single ground truth caption for a video  but there are many ground truths as a single video clip can have very many truthful descriptions of the video content depending on perspective, recent history, domain familiarity, even emotional state.  In such an environment, asking a user to caption a video, even asking several users to caption the same video, and then measuring automatically-generated captions against these, is not robust or reliable science.  What we are trying to do here is to measure the accuracy of automatically-generated video captions, some of which  may be fully accurate and correct while some others, will be inaccurate because the caption generation techniques are not perfect. So instead of measuring the generated captions against some groundtruth, we bring human assessment into the process and we draw on parallels with human evaluation in machine translation, which we describe in the next section.

\section*{Human Evaluation in Machine Translation}
\label{mtsection}

Evaluation in many areas of natural language processing (NLP) takes inspiration from Machine Translation evaluation,
including the tasks of  automatic summarization \cite{Graham:15} and grammatical error correction \cite{Gleu}.
Evaluation in Machine Translation (MT) commonly takes the form of comparison of 
automatic metric scores,
such as BLEU \cite{bleu}. Here, system performance is measured as the geometric mean
of matching proportions of n-gram counts between
the MT output with a human produced reference translation, in addition to a brevity penalty. 
However, automatic MT evaluation metrics are known to provide a less than perfect substitute 
for human assessment, as under some circumstances, it has been shown that an improvement
in BLEU is not sufficient to reflect a genuine improvement in translation quality, and 
in other circumstances that it is not necessary to improve BLEU in order to achieve a 
noticeable improvement \cite{callisonburchetal:06}.
Given the vast number of possible ways of comparing an MT output translation
with a human-produced reference translation, meta-evaluation of metrics is required to  
determine which metrics provide the most valid substitute for human
assessment. Such meta-evaluation commonly takes the form of the degree to 
which metrics scores correlate with human assessment.
In MT, the stronger the correlation of a metric with human assessment, the better
the metric is considered to be \cite{WMT16Metrics}.

To this end, the main benchmark in MT is the Workshop on Statistical Machine Translation (WMT),\footnote{The workshop has recently changed status to a conference but
maintains the original acronym.} where upwards of 150 world-leading MT systems 
annually participate in an evaluation campaign, and where the official results comprise 
human evaluation of systems \cite{WMT16}. In WMT, a large-scale effort involving  
volunteer human assessors and crowd-sourced workers is carried out across
several language pairs.
This human evaluation not only produces official results of the translation shared
task but also provides a gold standard for evaluation of newly proposed automatic
evaluation metrics.

Recent developments in MT have seen the development of new human assessment 
methodologies, one of which has  been adopted as the official method of
evaluation at WMT, Direct Assessment (DA).
DA involves direct estimation of the absolute quality of a given translation, in isolation from
other outputs, to avoid bias introduced when translations produced by differently performing
systems are compared more often to very high or low quality output, such as the positive bias 
known to be introduced when 
translations belonging to a system are compared more often to low quality 
output \cite{Bojaretal:11}.

In the case of MT, since genuinely bilingual human assessors are difficult to source, DA 
is structured as a monolingual task, where the human evaluator is required to compare
the meaning of the MT output with a human-produced reference translation, working within the same language.
Assessment scores are collected on a 0--100 rating scale, which facilitates comparison of
system performance based on human assessment score distributions. In addition this also allows
high-quality crowd-sourcing via quality control mechanisms based on significance testing of the
score distributions provided by workers.
The latter is highly important for carrying out assessment via the crowd, as due to the
anonymous nature of crowd-sourcing services, interference from workers attempting
to game the system and to maximize their profits is unfortunately unavoidable. Even when actively
rejecting assessment tasks submitted by dubious crowd-sourced workers, 
low quality task submission rates have been reported to be as high as 
between 38\% and 57\% \cite{WMT16}. 

The advantages of DA over previous methods of human evaluation are:
\begin{itemize}
\item Accurate quality control of crowd-sourcing \cite{Grahametal:14};
\item Absolute assessment of systems which allows measurement of longitudinal improvements in system performance \cite{Grahametal:14};
\item Results for both individual translations and for systems have been shown to be almost perfectly repeatable in self-replication experiments \cite{Grahametal:15, WMT16, Grahametal:17};
\item DA has been shown to be up to 10 times more effective at finding significant differences between competing systems compared to previous methodologies \cite{NLE:16};
\item DA has been shown to be  efficient and effective for tuning automatic evaluation metrics \cite{Maetal:17}.
\end{itemize}

\noindent 
In this paper we follow the Direct Assessment methodology used in evaluation of MT, and apply it to the monolingual task of comparing the quality of automatically generated video captions. In the next section we describe the data collection we used, the videos and the sets of descriptive captions generated for each,

\section*{Video-to-Text Track (VTT) in TRECVid}

The TREC Video Evaluation (TRECVid) benchmark has been active since 2003 in evaluating content-based video retrieval systems working on problems including but not limited to semantic indexing, video summarization, video copy detection, multimedia event detection, and ad-hoc video search. In 2016  a new showcase/pilot ``Video to Text Description'' (VTT) task \cite{awad2016trecvid} was proposed and launched within TRECVid motivated by many use case application scenarios which can greatly benefit from  such technology such as video summarization in the form of natural language,  facilitating the search and browsing of video archives using such descriptions, and describing videos to the blind, etc. In addition, learning video interpretation and temporal relations among events in a video will likely contribute to other computer vision tasks, such as prediction of future events from the video. In the following subsections we will review the data, task, evaluation and existing metrics used as well as the system results from participating groups.

\subsection*{Data and System Task}
A dataset of more than 30\,000 Twitter Vine videos has been collected automatically. Each video has a total duration of about 6s. A subset of 2\,000 of these videos was randomly selected and annotated manually, twice, by two different annotators. In total, 4 non-overlapping sets of 500 videos were given to 8 annotators to generate a total of 4\,000 text descriptions. Those 4\,000 text descriptions were split into 2 sets corresponding to the original 2\,000 videos. Annotators were asked to include and combine into 1 sentence, if appropriate and available, four facets of the video they are describing:
\begin{itemize}
\item Who is the video describing (e.g. concrete objects and beings, kinds of persons, animals, or things);
\item What are the objects and beings doing~? (generic actions, conditions/state or events);
\item Where is the video taken (e.g. locale, site, place, geographic location, architectural)
\item When is the video taken (e.g. time of day, season)
\end{itemize}

\noindent
After annotations were completed, an automatic filtering was applied to remove very short generic descriptions which resulted in ending up with only 1\,915 videos as the testing dataset.

The task set for participant groups was as follows: given a set of 1\,915 URLs of Vine videos and two sets (A and B) of text descriptions (each composed of 1\,915 sentences), participants were asked to develop systems  and submit results for automatically generating for each video URL, a 1-sentence text description  independently and without taking into consideration the existence of sets A and B.

\subsection*{Evaluation}
In total,  16 individual complete ``runs'' were submitted to the description generation subtask.
Evaluation of these was done automatically using  standard metrics from machine translation (MT) including METEOR* \cite{banerjee2005meteor} and BLEU* \cite{papineni2002bleu}.\footnote{We add * to metric names as the way in which they were applied to captions differs in some way to how scores are produced in a standard MT evaluation. Scores were computed on the segment-level for example} BLEU (bilingual evaluation understudy) is a metric used in MT and was one of the first metrics to achieve a high correlation with human judgments of quality. It is known to perform more poorly if it is used to evaluate the quality of individual sentence variations rather than sentence variations at a corpus level. In the VTT task the videos are independent thus there is no corpus to work from, so our expectations are lowered when it comes to evaluation by BLEU.  METEOR (Metric for Evaluation of Translation with explicit ORdering) is based on the harmonic mean of uni-gram or n-gram precision and recall, in terms of overlap between two input sentences. It redresses some of the shortfalls of BLEU such as better matching synonyms and stemming, though the two measures are used together in evaluating MT.   

Systems taking part in the VTT task were encouraged to take into consideration and use the four facets that annotators used as a guideline to generate their automated descriptions. 

In addition to using standard MT metrics, an experimental semantic similarity metric (STS) \cite{han2013umbc} was also applied. This automatic metric measures how semantically similar a submitted description is to the ground truth descriptions, either A or B.

\subsection*{Evaluation Results}

\begin{figure}[ht]
  \centering
  \includegraphics[width=0.95\columnwidth]{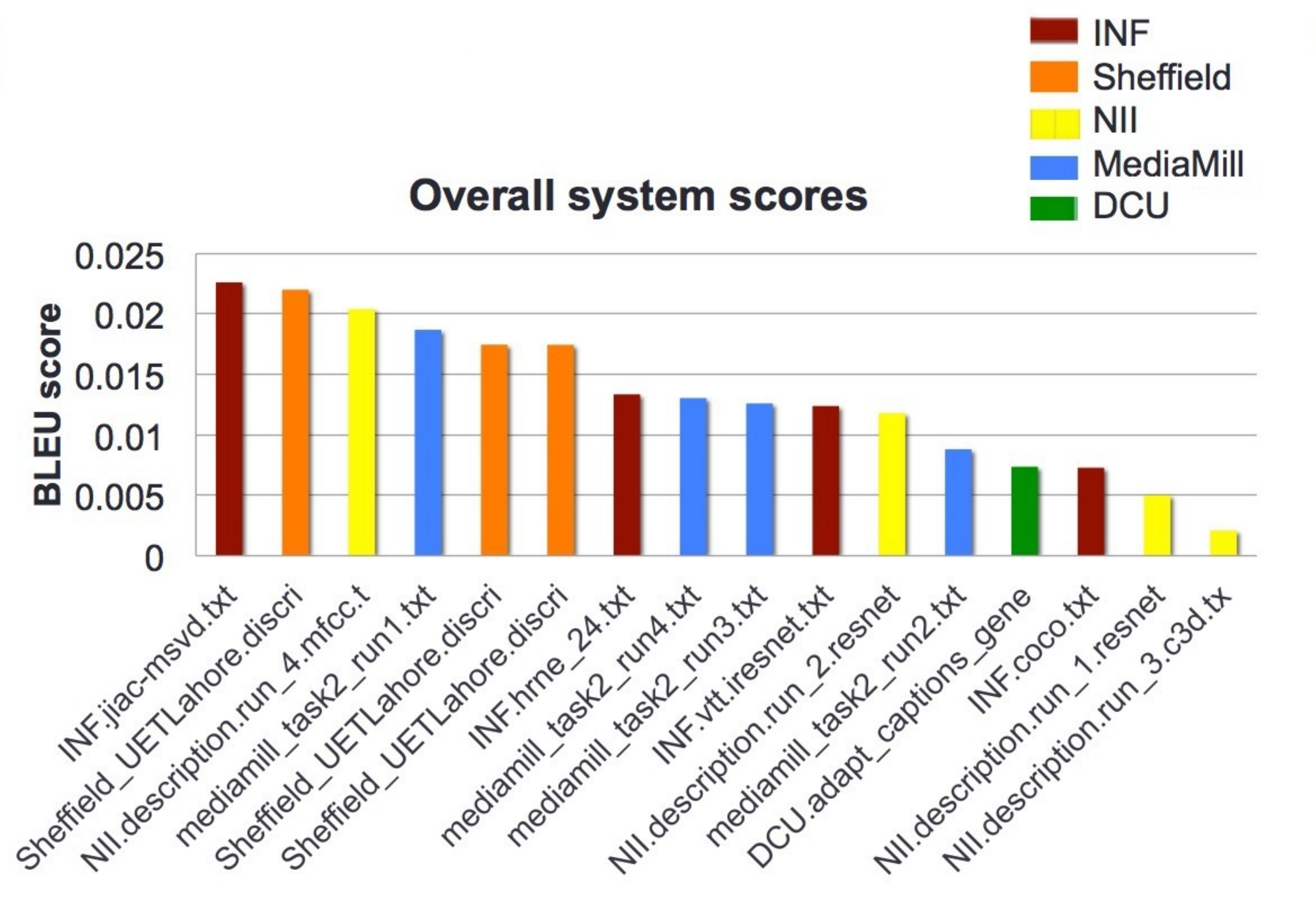}
  \caption{VTT: Results using the BLEU* metric (reproduced from \cite{awad2016trecvid})}
  \label{vtt.bleu}
\end{figure}

\begin{figure}[ht]
\centering
  \includegraphics[width=0.95\columnwidth]{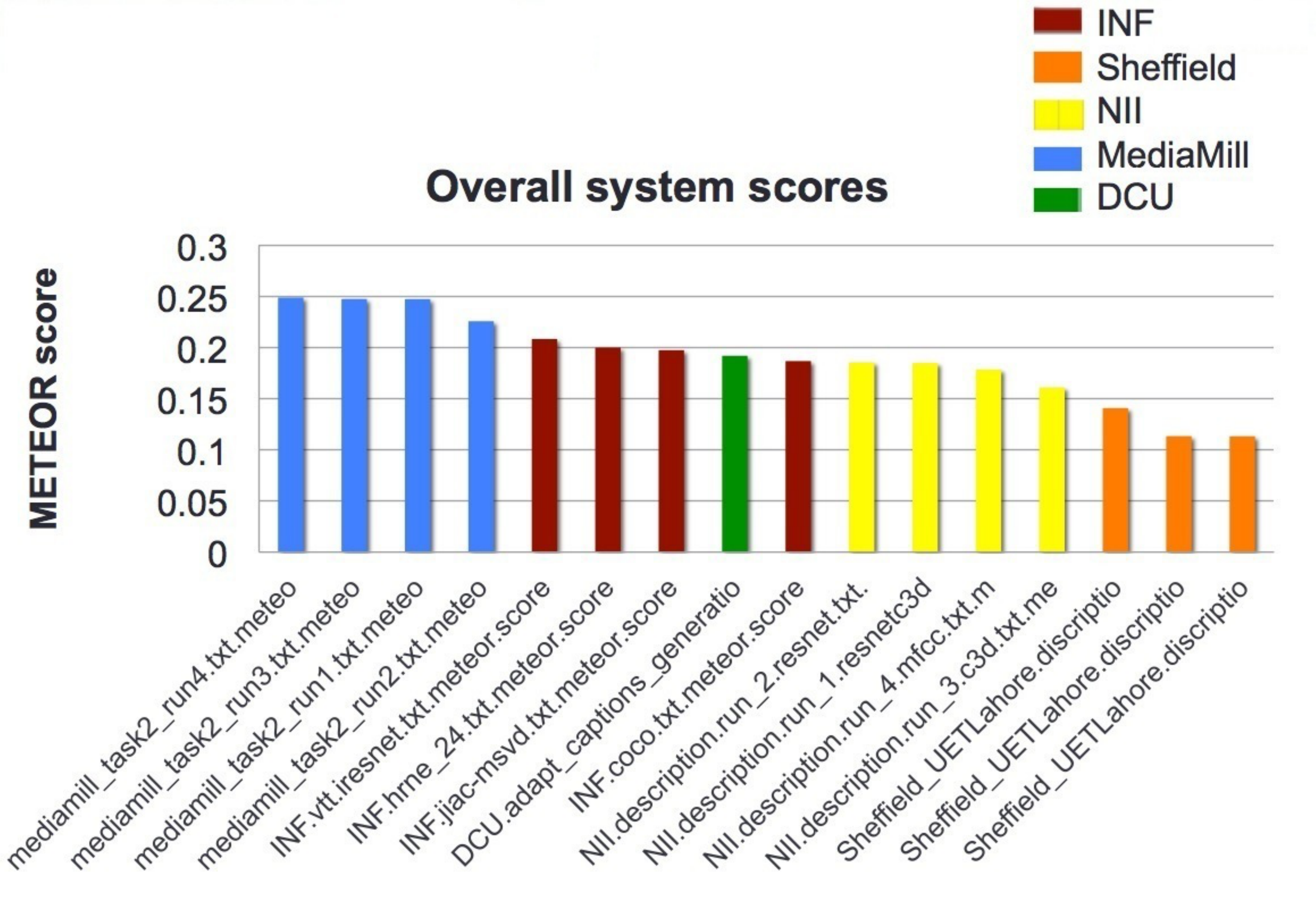}
  \caption{VTT: Results using the METEOR* metric  (reproduced from \cite{awad2016trecvid})}
  \label{vtt.meteor}
\end{figure}

Figs~\ref{vtt.bleu} and \ref{vtt.meteor} show the performance of the 16 runs submitted by the 5 participating groups using the BLEU* and METEOR* metrics respectively. 
The BLEU* results in Fig~\ref{vtt.bleu} are difficult to interpret because, for example, multiple results from single groups are scattered throughout the results list whereas one would expect results from a single site to cluster as each group usually submits only minor variations of its own system for generating captions. This may be due to the issues associated with using BLEU* for this task, as mentioned earlier.  The METEOR* results in Fig~\ref{vtt.meteor} show results for each group are indeed clustered by group and thus may be more reliable. However, for both BLEU* and METEOR*, trying to interpret the absolute values of the system scores is impossible so their real value is in comparison only.

\begin{figure}[ht]
  \centering
  \includegraphics[height=3.5in,width=0.9\columnwidth]{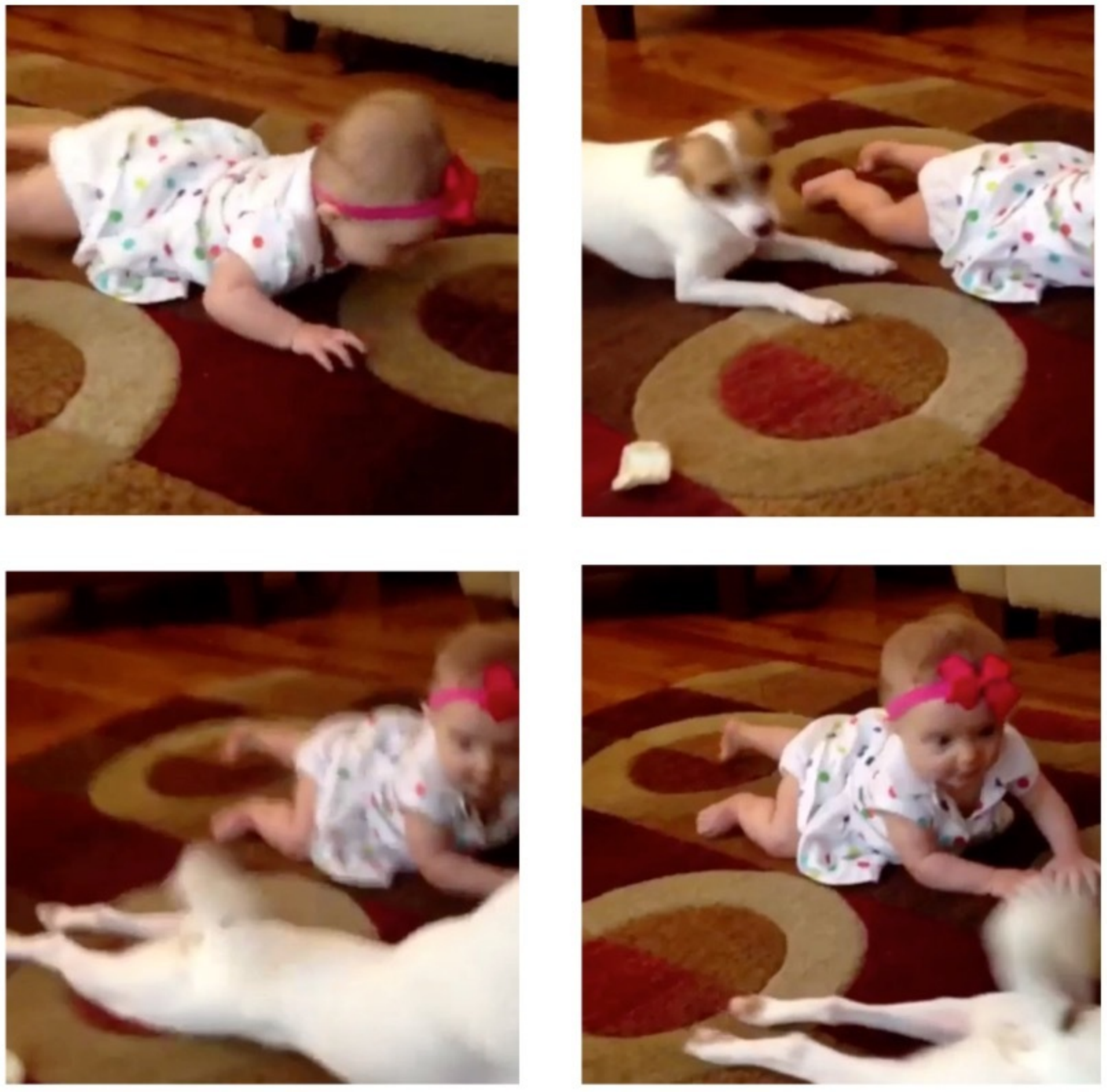}
  \caption{VTT: Sample video shown as a series of keyframes}
  \label{vtt.sample.frame}
\end{figure}

In order to give the reader some insight into the descriptive captions actually generated, Fig~\ref{vtt.sample.frame} shows a series of keyframes from one of the videos where a baby crawls forward left to right across what appears to be a living room carpet, the camera zooms out to reveal a dog behind the baby and the dog does indeed mimic the way the baby crawls with its hind legs trailing behind it.  

Below are the submitted captions for this video from across the groups (there are duplicate captions among the 16 submissions). 

\begin{itemize}
\item a girl is playing with a baby
\item a little girl is playing with a dog
\item a man is playing with a woman in a room
\item a woman is playing with a baby
\item a man is playing a video game and singing
\item a man is talking to a car
\item A toddler and a dog
\end{itemize}

What this shows is that there are good systems that do well, and others that do not do well in terms of the captions that they generate. Similarly there are videos which are easier to caption than others, and each approach does well on some videos and badly on others, but not consistently so. For detailed information about the approaches and results, the reader should see the various group workshop reports from the  5 participating groups and these are MediaMill from the University of Amsterdam \cite{snoek2016University}, the Informedia team from Carnegie Mellon University (INF) \cite{chen2016TRECVID}, Dublin City University (DCU) \cite{marsden2016dublin}, the National Institute for Informatics, Japan (NII) \cite{le2016nii}, and the University of Sheffield \cite{wahla2016the}.

\subsection*{Semantic Similarity Among Captions}

\begin{figure}[ht]
  \centering
  \includegraphics[width=0.9 \columnwidth]{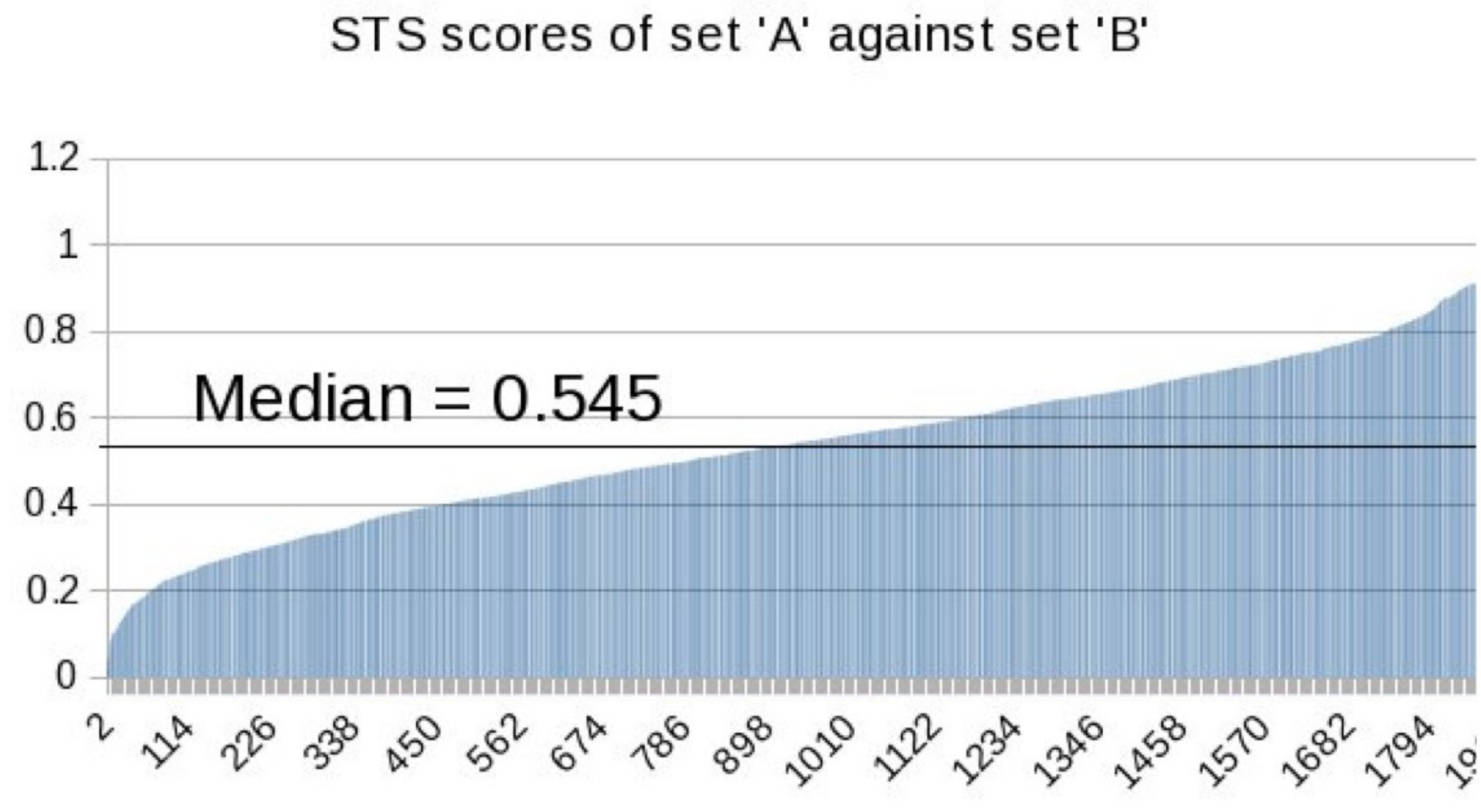}
  \caption{VTT: STS scores between the two reference ground truth sets (reproduced from \cite{awad2016trecvid})}
  \label{vtt.sts}
\end{figure}

In addition to BLEU* and METEOR* metrics, a semantics-based metric was also used in the VTT evaluation. 
This was the UMBC EBIQUITY-CORE metric \cite{han2013umbc} developed for the 2013 Semantic Textual Similarity (STS) task which uses  word similarity boosted by the use of WordNet which captures word relationships.
%
Fig~\ref{vtt.sts} shows the values of the STS metric where captions A and B for each of the 1\,915 videos in the evaluation are measured against each other. One would expect that the A and B captions would be semantically similar, perhaps even identical and so we would hope for a large number of the 1\,915 similarity measures to be at, or close to, a value of 1.0.  

\noindent
Instead, as Fig~\ref{vtt.sts} illustrates, the median similarity between pairs of captions (A and B) for each of the videos is only 0.545 with a disappointing tail-off of similarities close to a value of 0. That tells us that either the A and B annotators got things terribly wrong, 
or the STS measure has difficulty measuring similarity across just a short video caption, or the vocabulary used in the captioning creates difficulties for the STS computation.  Whatever the reason, STS similarity values would not add much value to interpreting the absolute performance of submitted runs though there might be some interesting insights gained from comparing relative performances.

\begin{figure}[ht]
\begin{center}
\includegraphics[width=1.0\textwidth]{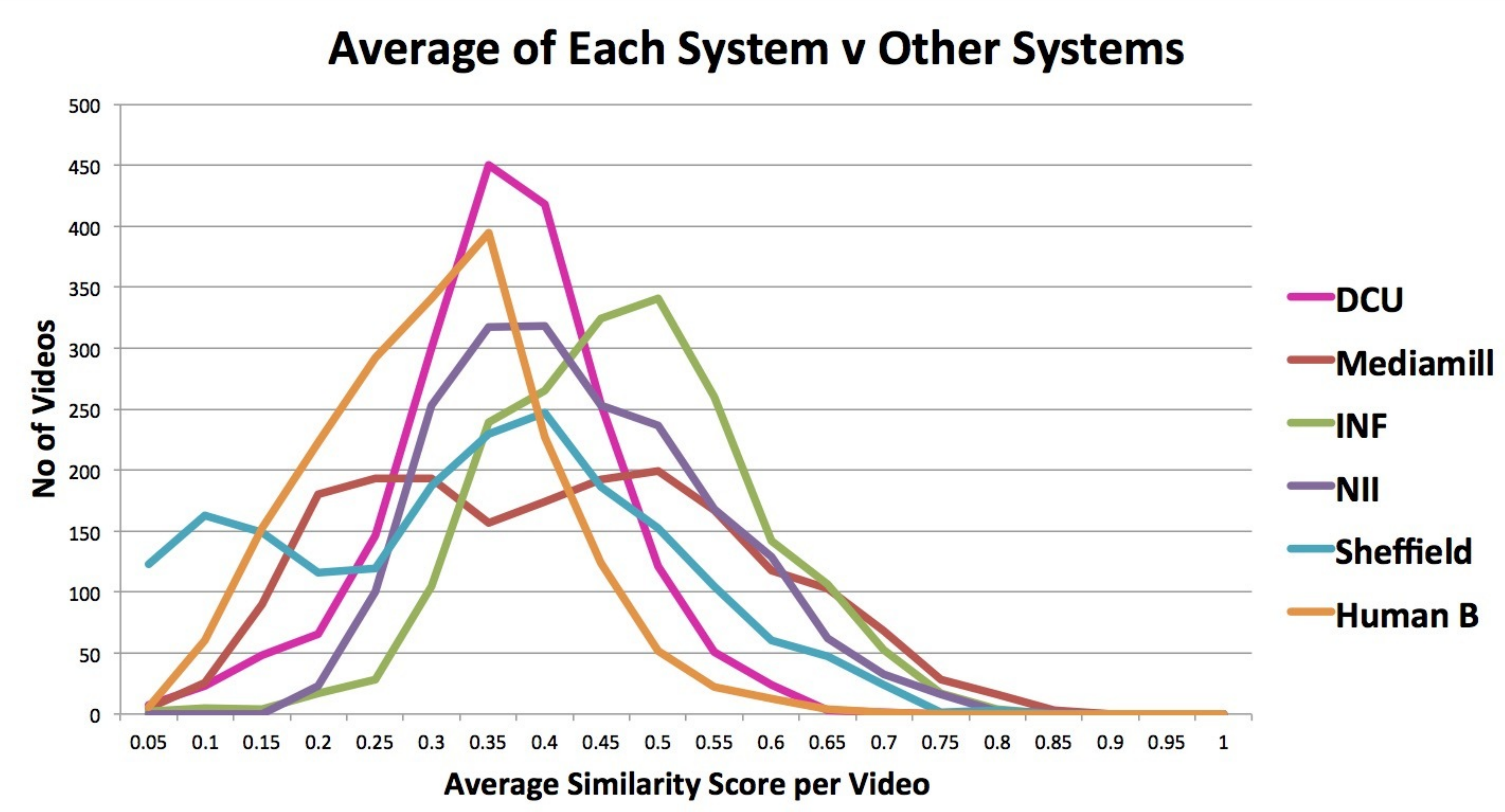}
\caption{TRECVid VTT task comparing participant performance and human annotation using STS inter-system similarities}
\label{fig:sts-sim}
\end{center}
\end{figure}

We used an API to the UMBC STS semantic similarity score\footnote{Available at \url{http://swoogle.umbc.edu/SimService/}} to calculate semantic similarities among submitted captions.
For the best-performing submitted run from each of the 5 groups (according to the other measures), for each video plus one of the sets of human annotations, we calculated the $6 \times 6$ STS similarities allowing us to see how semantically ``close'' or how ``far'' each of the submitted runs and the human annotation is to the others.  Averaged over the 1\,915 videos, this is shown in Figure~\ref{fig:sts-sim}

\begin{table}[ht]
\centering
\begin{tabular}{cc}
\toprule
Group & Average STS to other Groups \\
\midrule
INF & 0.446 \\
NII  & 0.405 \\
MediaMill & 0.384 \\
DCU & 0.338 \\
Sheffield & 0.308 \\
Human-b & 0.279 \\
\bottomrule
\\
\end{tabular}
\caption{Average STS Similarities from each system to all others}
\label{tab:STS-averages}
\end{table}

\noindent
What we  observe from this analysis is that the INF  group from Carnegie Mellon generates captions which are closer to the captions generated by the others including the human than do the groups from NII, MediaMill, DCU, Sheffield, or even the human annotation, in that order. This is visible from the graph in Figure~\ref{fig:sts-sim} where  the ``bulge'' for each of the systems is higher and closer to a similarity score of 1.0.  The line of the group from Sheffield shows an interesting kink at the very low end of the similarity scores and we interpret this as Sheffield generating some captions for some videos which are very inaccurate and thus quite dissimilar to any of the captions generated by any of the other systems or by the human annotator. From Table~\ref{tab:STS-averages} we see that the human annotations are more dissimilar to the others than any of the automatic captions. We believe this is due to the way human captions were generated where the annotaors were asked to include coverage of different facets of videos -- who (is present), where (did the video take place), when (was the video taken), etc.  This resulted in very structured annotations, almost formulaic in nature compared to the those generated by the automatic systems.

What this analysis means that if there was a {\em wisdom of the crowds} type of evaluation, where the ``best'' performing system was the one that was closed to all the others, and had fewest outliers, then the INF submission would be best. However this is evaluating caption performance from within the set of submitted runs rather than externalising the evaluation process, and this says nothing about absolute, only comparative performance.

\subsection*{Summary and Observations}

The first observation to make about the VTT task is that there was good participation from among TRECVid groups and that there are submitted captions with impressive results. Not all generated captions are correct or impressive, but there are enough good ones to be encouraged, meaning that proper evaluation for this is really now needed. 

In terms of metrics used, METEOR* scores are higher than BLEU*, and in retrospect the CIDEr metric (Consensus-based Image Description Evaluation) \cite{vedantam2015cider} should also have been used. As can be seen in their TRECVid workshop papers, some participating groups did include this metric in their write-ups. 
The STS semantic similarity metric is also useful but only insofar as it allows {\em comparative} performance across participating groups but such analysis does not offer any possibility for longitudinally tracking any progress over multiple iterations of the VTT task.

One aspect that became apparent as we looked at the approaches taken by different participants \cite{marsden2016dublin} is that there are lots of available training sets for this task, including MSR-VTT, MS-COCO, Place2, ImageNet, YouTube2Text, and MS-VD. Some of these even have manual ground truth captions generated with Mechanical Turk such as the MSR-VTT-10k dataset \cite{xu2016msr} which has 10\,000 videos, is 41.2 h in duration  and has 20 annotations for each video.
This provides a rich landscape for those wishing to use machine learning in all its various forms within the VTT task and participants in VTT have used all of these at some point.

\section*{Evaluating Video Captions}

Due to the many advantages of Direct Assessment for evaluation of MT systems outlined earlier,
we investigate the possibility of adapting DA human assessment to evaluation of
video captioning. 
In DA, the standard set-up for MT is to present  human assessors with a human-produced
reference
translation and an MT output in the same language and ask them to rate how well the latter expresses the
meaning of the former. In DA for 
MT, it is necessary to employ a human-produced reference translation instead of 
the original source language input, to avoid requiring bilingual human assessors.
In the case of video captioning, substitution of the video with a human-produced caption
would be risky, due to the ambiguity involved in a human coming up with a caption for
a video, as there 
are many correct but distinct captions for a single video,  the meaning of which may not 
directly correspond to one another, unlike reference translations in MT. This risk is clearly illustrated in Figure~\ref{vtt.sts} shown earlier where we see that even two manually created captions can't agree with each other when using the STS measure to assess.

Fortunately, however, substitution of the video with a human-produced caption is not only risky,
it is also unnecessary, as the video can itself be simply included in the evaluation and viewed by the human assessors before they rate an automatically generated caption. 
Subsequently, we ask human assessors to 
firstly watch the video before reading the caption, and then assess how adequately the caption (to be
evaluated) describes what took place in the video.  Assessors are asked to rate the caption quality on a scale of 1 to 100.

\begin{figure}[ht]
\begin{center}
\includegraphics[width=0.85\textwidth]{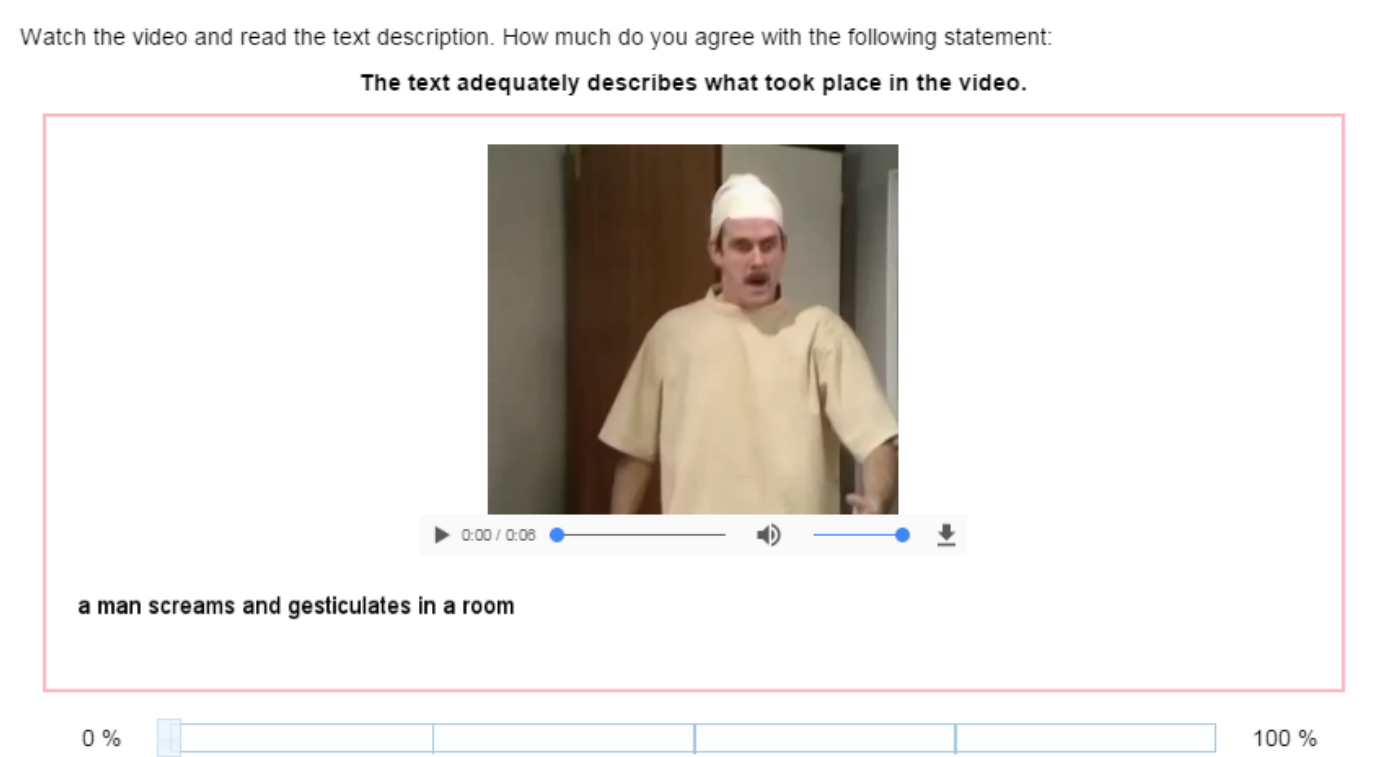}
\caption{Screen shot of Direct Assessment of Video Captioning on Mechanical Turk }
\label{screenshot}
\end{center}
\end{figure}

Figure \ref{screenshot} shows a screen shot of our DA human assessment set-up for video 
captioning as shown to workers on the crowd-sourcing service Amazon's Mechanical
Turk.\footnote{\url{https://www.mturk.com}}

\subsection*{Quality Controlling the Crowd}

As in DA for MT, we devise a mechanism for distinguishing genuine and diligent human assessors of
video captions from those attempting to game the service or those
who are simply not applying sufficient human intelligence to the task,
by establishing the consistency of individual workers at scoring high and low quality captions
for videos in the test set.

We therefore include in the evaluation of systems, an additional system comprising
a set of human annotations for the video captioning test set, (A), that will act as a set of
captions that we expect to receive high scores from workers.
We then create a corresponding set of low quality captions for those videos 
by automatically degrading each the human caption in (A).
Captions are automatically degraded by randomly selecting a non-initial/non-final
sequence of words within the original human caption and replacing it with a sequence
of words randomly selected from the human caption of a different video. In this way,
the sequence of words being substituted is a fluent sequence of words, making it difficult
to spot without reading the entire caption, to help avoid further gaming approaches.
The length of the phrase to be replaced is determined by the number of words in the
caption to be degraded, as shown in Table~\ref{tab:subs}:
\begin{table}
\begin{center}
\begin{tabular}{cc}
\toprule
Caption & \# Words Replaced \\
Length (N) & in Caption \\
\midrule
1        & 1 \\
2--5     & 2 \\
6--8     & 3 \\
9--15    & 4 \\
16--20   & 5 \\
$>$20      & $\lfloor$ N/4 $\rfloor$ \\
\bottomrule
\\
\end{tabular}
\caption{Substitution rules when degrading captions}
\label{tab:subs}
\end{center}
\end{table}

As in DA MT evaluation, Human Intelligence Tasks (HITs) are structured as 
100-captions per HIT (or 100-translation HITs for MT) as this allows sufficient 
pairs of high quality and low quality captions 
to be hidden within HITs and collected from each worker who participates. 
A minimum of 10 such pairs of high and low quality captions are therefore collected
from each worker and a paired significance test is then applied to the score distributions
of high and low quality captions for each individual worker. The p-value produced in the test
is employed as an estimate of worker reliability to score low quality captions lower
than the corresponding high quality captions, with a lower p-value indicating a more
consistent human assessor.

\subsection*{Evaluation}

In order to investigate the viability of our new video captioning human evaluation
methodology, we carried out an evaluation of systems participating in TRECVid in 2016, described earlier in this paper.  
We include in the evaluation
a single submission for each team that participated in the competition, selected from the runs they originally submitted.

Crowd-sourced workers were paid at the same rate as in our MT experiments, at 
0.99 USD per 100-caption HIT, and including the 20\% Mechanical Turk fee this resulted
in a total cost of 110 USD for the evaluation of five systems, or 22 USD on average per system.\footnote{The cost of adding additional systems increases at a linear rate.} 
A total of 90 workers completed at least
a single 100-video HIT on Mechanical Turk with a relatively high proportion of those, 72 (80\%),
meeting our quality control requirement of a p-value less than 0.05 for significance
of differences in scores they assigned to high quality and low quality translations.
In addition to including high and low quality pairs of translations within HITs, we also
include exact repeats of the same video and caption sampled across all systems included
in the evaluation, in order to check that workers who can distinguish between the quality
of high and low quality captions, also consistently score repeats of the same captions.
100\% of workers who passed quality control also showed no significant difference in 
score distributions for repeat assessment of the same caption.

Final scores for systems are computed by firstly standardizing the raw scores 
provided by
each worker (to $z$ scores which are the number of standard deviations from the mean a data point is) according to that individual's mean and standard deviation score overall,
as this allows any possible bias introduced by, for example, an overly harsh worker
evaluating a higher number of captions from any system. Scores provided are then
averaged for each caption in the test set, as some captions will have been assessed
once and others multiple times, and this avoids final scores inadvertently
being weighted by the number of times a given caption was assessed. The final score
for a given system is then simply the average of the scores attributed to its captions.

\begin{figure}
\begin{center}
\includegraphics[width=0.35\textwidth]{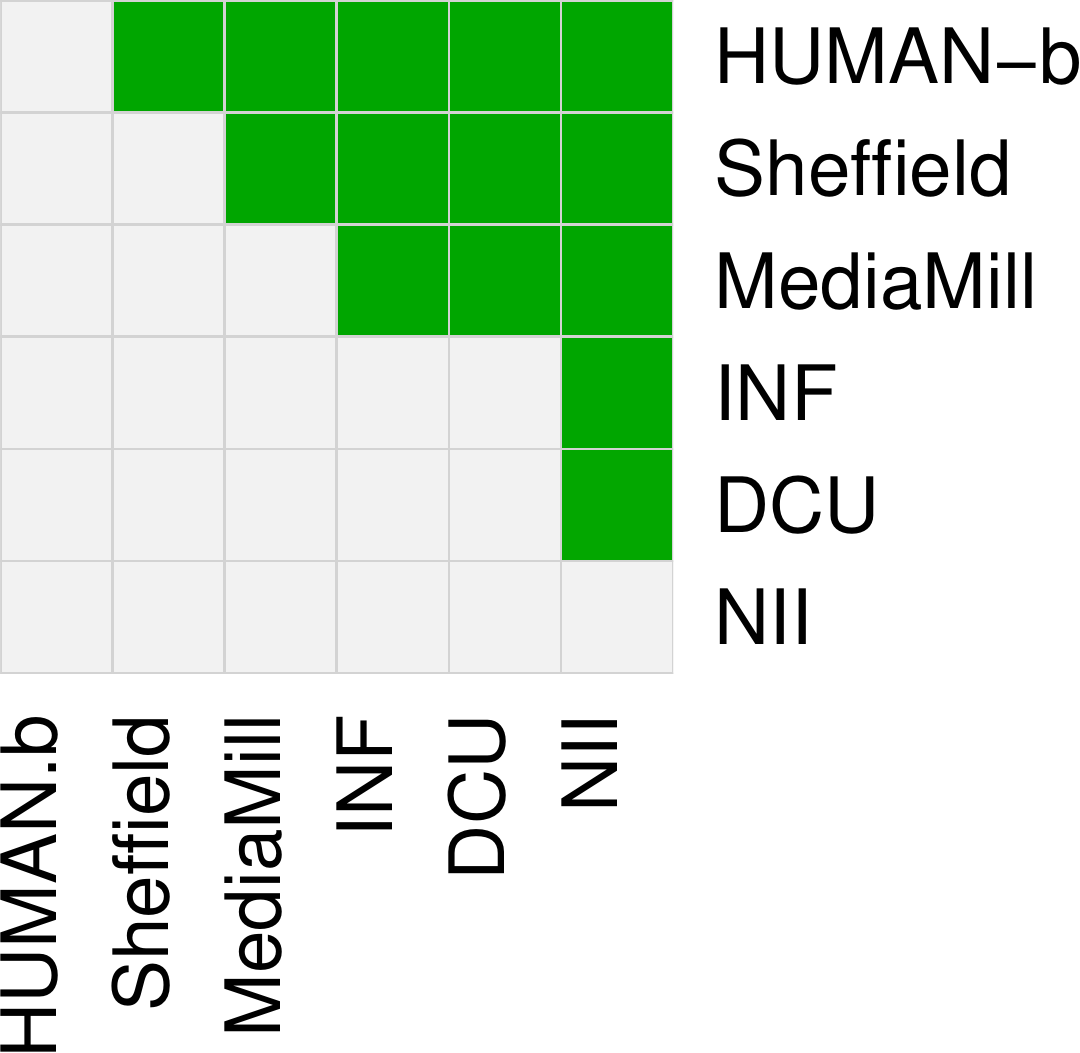}
\caption{Significance test results for video captioning systems originally participating in
TRECVid 2016 VTT task according to Wilcoxon rank-sum test based
on $z$ score distributions for systems. A green
cell in a given row signifies a significant win for the system
in that row over the system in that column. Human-b is a set of 
video captions produced by human annotators.}\label{sig}
\end{center}
\end{figure}

Table \ref{results} shows the raw average DA scores for each system included in 
the evaluation as well as the set of human captions, one of the sets of ground truths 
provided by NIST, included as a hidden system, while Figure \ref{sig} shows significance 
test results for differences in (the more fair) $z$ score distributions for systems according to 
Wilcoxon rank-sum test. As expected, the human captions, Human-b,
achieve the highest score overall, at 88.2\%, and including such a \emph{human} system allows
an estimation of the score that could be expected from a system that has effectively solved the video captioning problem. Scores for the actual systems range from 37.7 to 66.2 \%, showing the top system to be still some distance from human annotation quality. In terms of statistical significance, we see an almost absolute ranking of systems, with only a single pair of systems (INF \& DCU) tied with no significant difference in their scores. 
\begin{table}
\begin{center}
\begin{tabular}{lcccc}
\toprule
System 		& raw  (\%) 	& $z$        			 & $n$ \\
\midrule
Human-b	                   & 88.2 			& \SD0.895 	  & \SN940 \\ 
Sheffield 	 & 66.2 		  &  \SD0.303   & 1,301 \\ 
MediaMill 	           & 49.0 			& $-$0.129  	  & 1,292 \\ 
INF                  & 42.9 			& $-$0.278 	  & 1,338 \\ 
DCU  	           & 41.3 			& $-$0.314 	   &  1,302 \\ 
NII                & 37.7 			& $-$0.423 	  & 1,347 \\ 
\bottomrule
\end{tabular}
\caption{Direct Assessment human evaluation results of systems originally participating in the 
TRECVid 2016 VTT task; raw scores are average scores for systems;
$z$ scores are average standardized scores for systems;
Human-b comprises one of the set of video captions produced by human
annotators. }\label{results}
\end{center}
\end{table}

In order to investigate the reliability of the results produced by the new DA
video captioning evaluation, we re-ran the evaluation in an entirely separate and repeated data
collection/assessment  on Mechanical Turk  and compared results. Table \ref{cor}
shows the correlation between raw average scores for systems in both the original and the repeated
runs, as well as the correlation between average $z$ scores for systems, the
official method of system ranking we recommend.
The very high correlation between scores for systems confirms that the
method of evaluation is reliable, producing system rankings that are almost perfectly 
reproducible for evaluation of video captioning.

\begin{table}
\begin{center}
\begin{tabular}{lcc}
\toprule
  	    & raw & $z$ \\
\midrule
$r$     & 0.995 & 0.997 \\
\bottomrule
\end{tabular}
\caption{Pearson correlation ($r$) of scores for systems produced in two separate
data collection runs on Mechanical Turk.}\label{cor}
\end{center}
\end{table}

\subsection*{Metric Correlation with Human Assessment}

\begin{figure}
\centering
\begin{tabular}{c}
\begin{tabular}{lll}
(a)  & (b) \\
 \includegraphics[width=0.4\columnwidth]{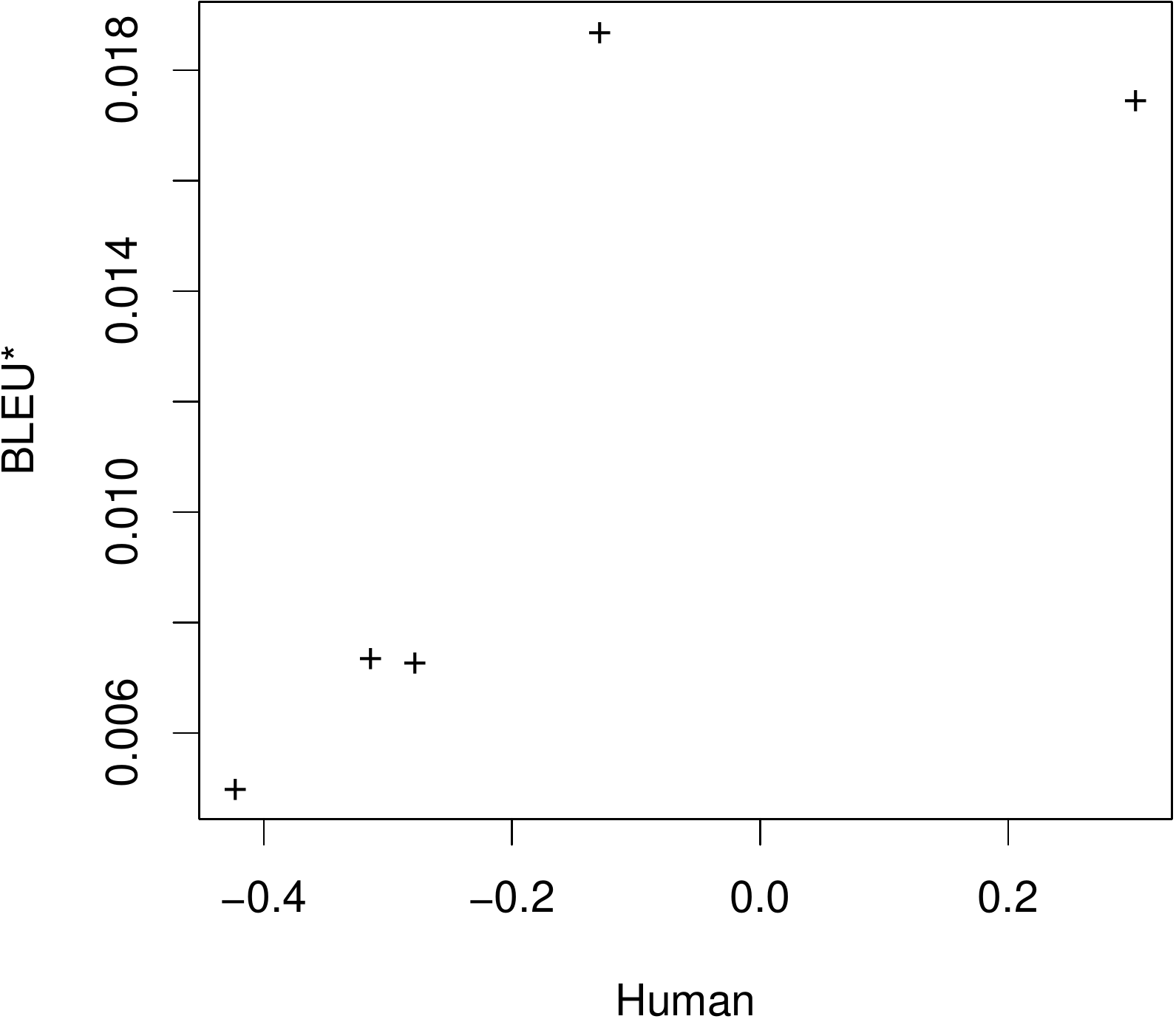} & \includegraphics[width=0.4\columnwidth]{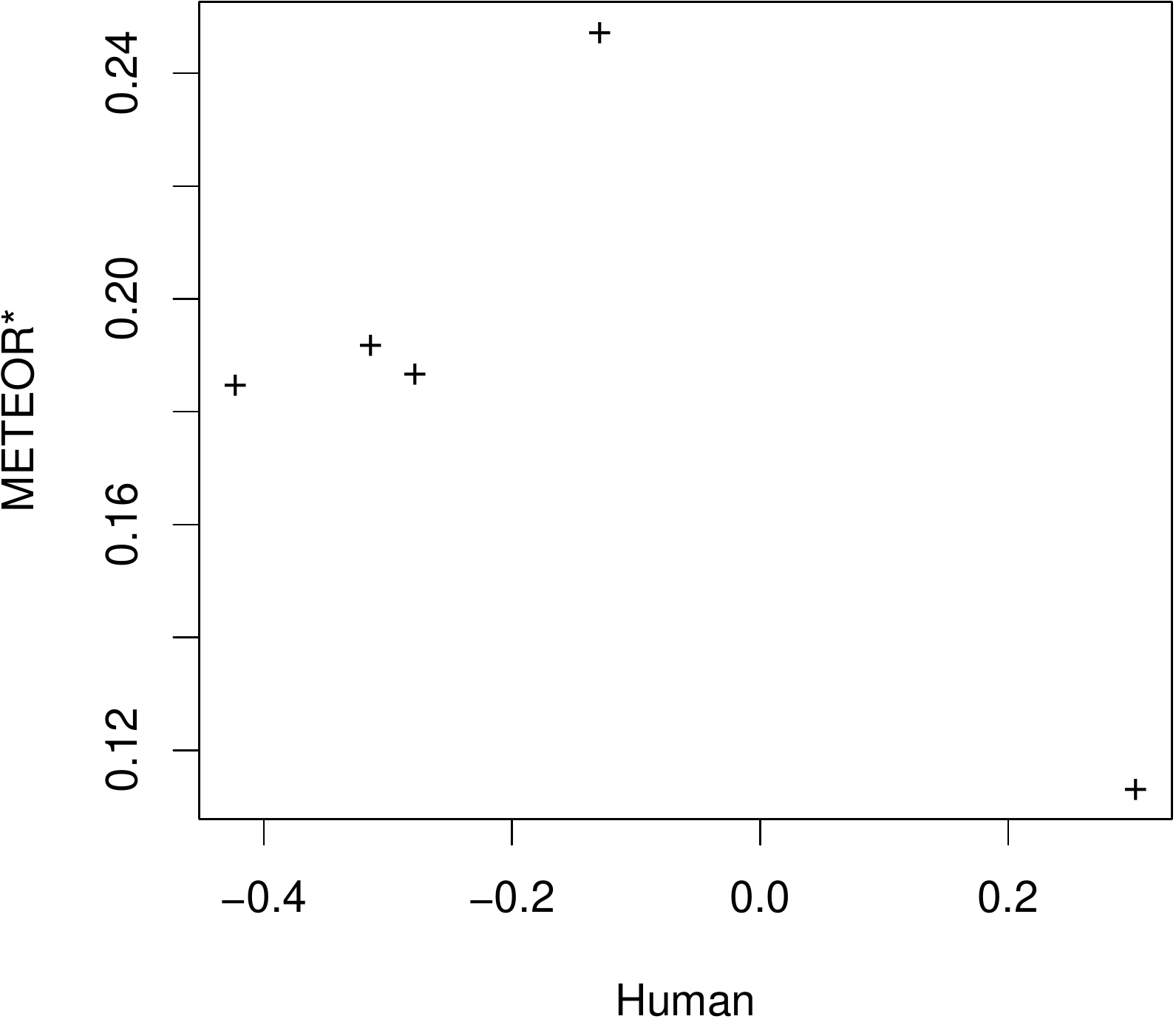}
 \\
\end{tabular} \\
\begin{tabular}{lll}
(c) \\
 \includegraphics[width=0.4\columnwidth]{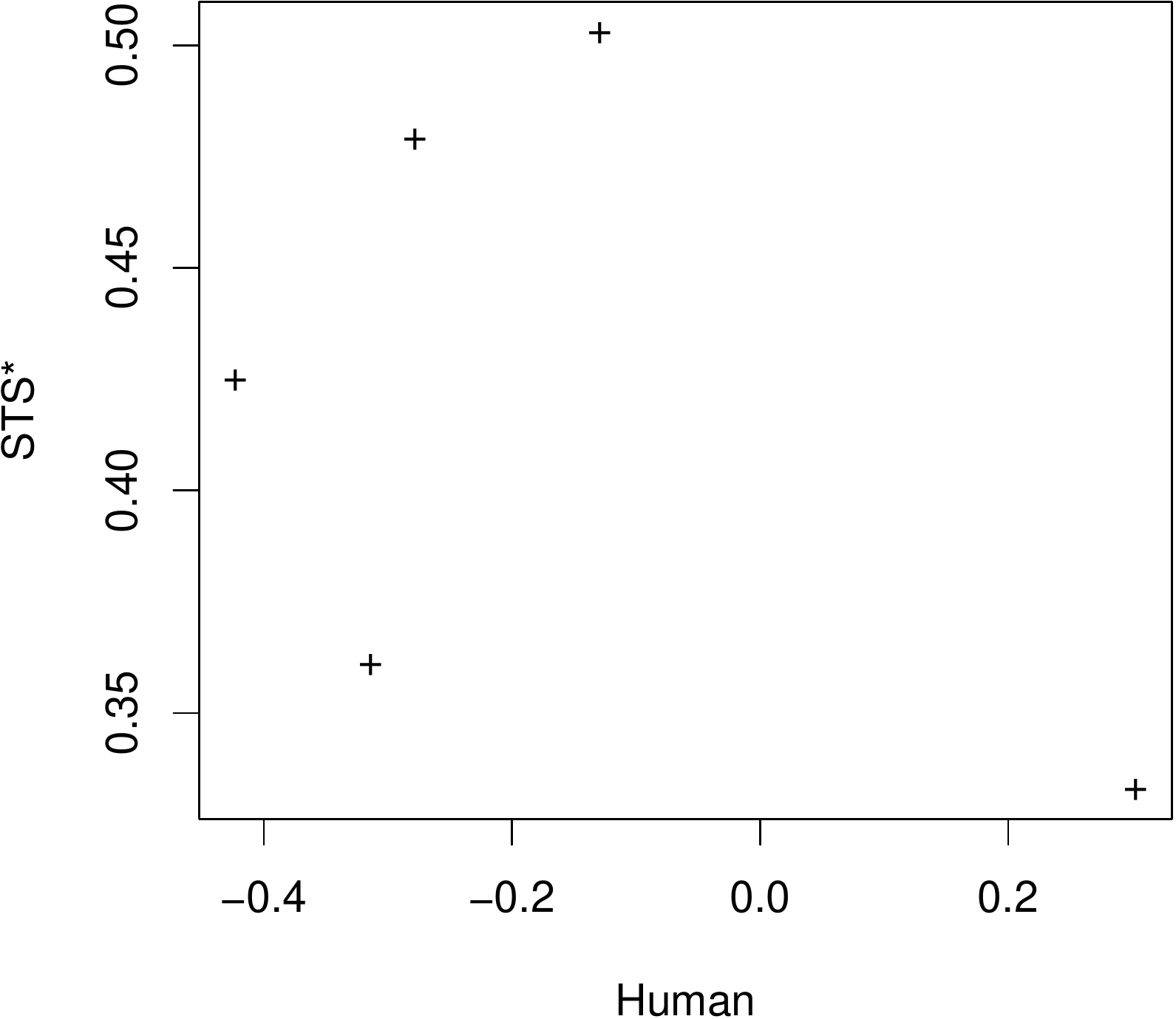} \\
\end{tabular} \\
\end{tabular}
\caption{Correlation between automatic metric scores and human scoring of systems.}\label{metric-plotsa}
\end{figure}

As mentioned previously, in MT the use of automatic metrics is
validated by how well a given metric correlates with human assessment. Since
now have human assessment of captions for systems participating
in TRECVid 2016 VTT, we can therefore examine how well metric scores
used to produce the official results in last year's benchmark correlate with human
assessment.
Table \ref{metric-r} shows the degree to which BLEU*, METEOR* and STS, as applied to captions in last year's VTT task, correlate with human assessment.
\begin{table}
\centering
\begin{tabular}{lcccc}
\toprule
    & BLEU* & METEOR*  & STS & $n$ \\
\midrule
$r$ &  0.800 & $-$0.602  & $-$0.444 & 5 \\
\bottomrule
\end{tabular}
\caption{Correlation of BLEU*, METEOR* and STS scores for submissions participating in 
in TRECVid 2016 VTT task with human assessment.}\label{metric-r}
\end{table}
To provide more detail of how metric and human scores correspond to one another, Figures \ref{metric-plotsa}(a),  \ref{metric-plotsa}(b) and \ref{metric-plotsa}(c) show respective scatter-plots of BLEU*, METEOR and STS scores and human assessment of systems. 

As can be seen in Fig~\ref{metric-plotsa}(a),  BLEU* scores correspond reasonably well to human evaluation, with the main disagreement taking place for the top two systems, Sheffield and MediaMill, the former outperforming the latter according to human evaluation. In contrast BLEU* scores incorrectly report a higher score for MediaMill, although only marginally so.
Fig~\ref{metric-plotsa}(b), on the other hand, shows substantially lower agreement between METEOR* scores and human assessment compared to BLEU*, where the system scored highest according to human judges and significantly outperforming all other systems, Sheffield, is ranked worst according to this metric. This disagreement in fact results in a somewhat disconcerting \emph{negative} correlation of METEOR* scores with human assessment of $-$0.602 for video captioning, as shown in Table~\ref{metric-r}, highlighting the potential danger in blindly trusting  automatic metric scores and the importance of meta-evaluation by correlation with human assessment. Finally, STS scores also show low agreement with human assessment, as the scatter plot in Fig~\ref{metric-plotsa}(c) shows, where, again, the best system according to human judges receives the lowest of the five STS scores. 



It should be noted, however, that the correlations we report here may not provide an entirely reliable evaluation of metrics for VTT. Both the human and metric-based evaluations reported are only based on a small sample of five VTT systems, and therefore correlation point estimates are not highly reliable. Significance tests, recommended for evaluation of MT metrics and also suitable for VTT metric evaluation, Williams test \cite{williams-test, GrahamBaldwin:14}, do however indicate that even at this low sample size BLEU*'s correlation with human assessment is significantly higher than that of both METEOR* (at $p<$ 0.01) and STS (at $p < $ 0.05).

\section*{Conclusions and Future Plans}

In this paper we have introduced direct assessment, a method for incorporating human judgments into an evaluation of automatic video captioning. Using data from the 2016 VTT track in TRECVid we have shown how this method is robust.  Even though direct assessment requires human judgments and ratings of how well a caption describes a video, this can be outsourced to a service like Mechanical Turk meaning it can be carried out in a relatively cost-effective manner.

One of the known problems with crowdsourced judgments as we use in direct assessment, is achieving consistency and quality across human  judges. We address this by automatically modifying candidate captions and degrading their quality and examining how the degraded captions are rated by the human judges. This allows us to apply quality control to the assessors meaning that direct assessment can be replicated consistently, something we demonstrated in the paper.

As part of the VTT track of TRECVid 2017, we expect a substantial increase in the number of participating groups with almost 30 signed up to participate at the time of writing.  Not all of these will submit runs but for those who do, each is asked to nominate a priority run which we will use, along with 5 independent sets of manual ground truth captions created by NIST, in our direct assessment method.  Having a greater number of candidate captions for each video will provide more reliable correlation point estimates for human and automatic metric scores. We also plan to provide a meta-evaluation of a range of applications of BLEU* and METEOR* to the evaluation of video captioning.

\section*{Acknowledgments}
The work reported in this paper has been supported by Science Foundation Ireland (\url{sfi.ie}) under the SFI Research Centres Programme co-funded under the European Regional Development Fund, grant numbers SFI/12/RC/2289 (Insight Centre) and 13/RC/2106 (ADAPT Centre for Digital Content Technology, \url{www.adaptcentre.ie}) at Dublin City University and by the NIST TRECVID workshop (\url{trecvid.nist.gov}) for running the VTT task and supporting the evaluation of systems.  We also thank Bronwyn Conroy for work done in generating STS evaluation graphs and Marc Ritter and his team for generating the manual video annotations.

\section*{Author Contributions}
Conceived and designed the experiments: YG, AS. Performed the experiments: YG, GA, AS. Analyzed the data: YG, AS. Contributed reagents/materials/analysis tools: AS, GA. Wrote the paper: YG, GA, AS.

\nolinenumbers

%
%
%


\bibliography{VTT-bibliography}

\begin{thebibliography}{10}

\bibitem{GUO201627}
Guo Y, Liu Y, Oerlemans A, Lao S, Wu S, Lew MS.
\newblock Deep learning for visual understanding: A review.
\newblock Neurocomputing. 2016;187:27 -- 48.
\newblock doi:{http://dx.doi.org/10.1016/j.neucom.2015.09.116}.

\bibitem{awad2016trecvida}
Awad G, Snoek CG, Smeaton AF, Qu{\'e}not G.
\newblock {TRECVid} Semantic Indexing of Video: A 6-Year Retrospective.
\newblock ITE Transactions on Media Technology and Applications.
  2016;4(3):187--208.

\bibitem{smeaton2006evaluation}
Smeaton AF, Over P, Kraaij W.
\newblock Evaluation campaigns and TRECVid.
\newblock In: Proceedings of the 8th ACM international workshop on Multimedia
  information retrieval. ACM; 2006. p. 321--330.

\bibitem{russakovsky2015imagenet}
Russakovsky O, Deng J, Su H, Krause J, Satheesh S, Ma S, et~al.
\newblock Imagenet large scale visual recognition challenge.
\newblock International Journal of Computer Vision. 2015;115(3):211--252.

\bibitem{muller2010experimental}
M{\"u}ller H, Clough P, Deselaers T, Caputo B, CLEF I.
\newblock Experimental evaluation in visual information retrieval.
\newblock The Information Retrieval Series. 2010;32.

\bibitem{Zafarani:2015:EWG:2783419.2666680}
Zafarani R, Liu H.
\newblock Evaluation Without Ground Truth in Social Media Research.
\newblock Commun ACM. 2015;58(6):54--60.
\newblock doi:{10.1145/2666680}.

\bibitem{kramer2014experimental}
Kramer AD, Guillory JE, Hancock JT.
\newblock Experimental evidence of massive-scale emotional contagion through
  social networks.
\newblock Proceedings of the National Academy of Sciences.
  2014;111(24):8788--8790.

\bibitem{Graham:15}
Graham Y.
\newblock Re-evaluating Automatic Summarization with BLEU and 192 Shades of
  ROUGE.
\newblock In: Proceedings of the 2015 Conference on Empirical Methods in
  Natural Language Processing. Lisbon, Portugal: Association for Computational
  Linguistics; 2015. p. 128--137.
\newblock Available from: \url{http://aclweb.org/anthology/D15-1013}.

\bibitem{Gleu}
Mutton A, Dras M, Wan S, Dale R.
\newblock GLEU: Automatic evaluation of sentence-level fluency.
\newblock In: Proceedings of the 45th Annual Meeting of the Association for
  Computational Linguistics. vol.~7. Association for Computational Linguistics;
  2007. p. 344--351.

\bibitem{bleu}
Papineni K, Roukos S, Ward T, Zhu WJ.
\newblock {BLEU}: A Method for Automatic Evaluation of Machine Translation.
\newblock IBM Research, Thomas J. Watson Research Center; 2001. RC22176
  (W0109-022).

\bibitem{callisonburchetal:06}
Callison-Burch C, Osborne M, Koehn P.
\newblock Re-evaluating the Role of BLEU in Machine Translation Research.
\newblock In: 11th Conference of the European Chapter of the Association for
  Computational Linguistics. Trento, Italy: Association for Computational
  Linguistics; 2006. p. 249--256.
\newblock Available from: \url{http://aclweb.org/anthology-new/E/E06/E06-1032}.

\bibitem{WMT16Metrics}
Bojar O, Graham Y, Kamran A, Stanojevi\'{c} M.
\newblock Results of the WMT16 Metrics Shared Task.
\newblock In: Proceedings of the First Conference on Machine Translation.
  Berlin, Germany: Association for Computational Linguistics; 2016. p.
  199--231.
\newblock Available from: \url{http://www.aclweb.org/anthology/W/W16/W16-2302}.

\bibitem{WMT16}
Bojar O, Chatterjee R, Federmann C, Graham Y, Haddow B, Huck M, et~al.
\newblock Findings of the 2016 Conference on Machine Translation.
\newblock In: Proceedings of the First Conference on Machine Translation.
  Berlin, Germany: Association for Computational Linguistics; 2016. p.
  131--198.
\newblock Available from: \url{http://www.aclweb.org/anthology/W/W16/W16-2301}.

\bibitem{Bojaretal:11}
Bojar O, Ercegov\v{c}evi\/{c} M, Popel M, Zaidan O.
\newblock A Grain of Salt for the {WMT} Manual Evaluation.
\newblock In: Proc. 6th Wkshp. Statistical Machine Translation. Edinburgh,
  Scotland: Association for Computational Linguistics; 2011. p. 1--11.

\bibitem{Grahametal:14}
Graham Y, Baldwin T, Moffat A, Zobel J.
\newblock Is Machine Translation Getting Better over Time~?
\newblock In: Proceedings of the 14th Conference of the European Chapter of the
  Association for Computational Linguistics. Gothenburg, Sweden: Association
  for Computational Linguistics; 2014. p. 443--451.
\newblock Available from: \url{http://www.aclweb.org/anthology/E14-1047}.

\bibitem{Grahametal:15}
Graham Y, Mathur N, Baldwin T.
\newblock Accurate Evaluation of Segment-level Machine Translation Metrics.
\newblock In: Proceedings of the 2015 Conference of the North American Chapter
  of the Association for Computational Linguistics Human Language Technologies.
  Denver, Colorado: Association for Computational Linguistics; 2015.

\bibitem{Grahametal:17}
Graham Y, Ma Q, Baldwin T, Liu Q, Parra C, Scarton C.
\newblock Improving Evaluation of Document-level MT Quality Estimation.
\newblock In: Proceedings of the 15th European Chapter of the Association for
  Computational Linguistics. Valencia, Spain: Association for Computational
  Linguistics; 2017.

\bibitem{NLE:16}
Graham Y, Baldwin T, Moffat A, Zobel J.
\newblock Can Machine Translation Systems be Evaluated by the Crowd Alone.
\newblock Natural Language Engineering. 2017;23:3--30.
\newblock doi:{10.1017/S1351324915000339}.

\bibitem{Maetal:17}
Ma Q, Graham Y, Wang S, Liu Q.
\newblock Blend: a Novel Combined MT Metric Based on Direct Assessment.
\newblock In: Proceedings of the Second Conference on Machine Translation.
  Copenhagen, Denmark: Association for Computational Linguistics;.

\bibitem{awad2016trecvid}
Awad G, Fiscus J, Michel M, Joy D, Kraaij W, Smeaton AF, et~al.
\newblock {TRECVid} 2016: Evaluating video search, video event detection,
  localization, and hyperlinking.
\newblock In: Proceedings of TRECVID. vol. 2016; 2016.

\bibitem{banerjee2005meteor}
Banerjee S, Lavie A.
\newblock METEOR: An {A}utomatic {M}etric for {MT} {E}valuation with {I}mproved
  {C}orrelation with {H}uman {J}udgments.
\newblock In: Proceedings of the ACL workshop on intrinsic and extrinsic
  evaluation measures for machine translation and/or summarization. vol.~29;
  2005. p. 65--72.

\bibitem{papineni2002bleu}
Papineni K, Roukos S, Ward T, Zhu WJ.
\newblock {BLEU}: {A} {M}ethod for {A}utomatic {E}valuation of {M}achine
  {T}ranslation.
\newblock In: Proceedings of the 40th Annual Meeting of the Association for
  Computational Linguistics. Association for Computational Linguistics; 2002.
  p. 311--318.

\bibitem{han2013umbc}
Han L, Kashyap A, Finin T, Mayfield J, Weese J.
\newblock {UMBC} {EBIQUITY-CORE}: {S}emantic {T}extual {S}imilarity {S}ystems.
\newblock In: Proceedings of the Second Joint Conference on Lexical and
  Computational Semantics. vol.~1. Association for Computational Linguistics;
  2013. p. 44--52.

\bibitem{snoek2016University}
Snoek CGM, Gavves E, Hussein N, Koelma DC, Smeulders AWM, Dong J, et~al.
\newblock University of Amsterdam and Renmin University at TRECVID 2016:
  Searching Video, Detecting Events and Describing Video.
\newblock In: TRECVID 2016 Workshop. Gaithersburg, MD, USA; 2016.

\bibitem{chen2016TRECVID}
Chen J, Pan P, Huang P, Fan H, Sun J, Yang Y, et~al.
\newblock {INF@TRECVID 2016: Video Caption Pilot Task}.
\newblock In: TRECVID 2016 Workshop. Gaithersburg, MD, USA; 2016.

\bibitem{marsden2016dublin}
Marsden M, Mohedano E, McGuinness K, Calafell A, Giro-i Nieto X, O'Connor NE,
  et~al.
\newblock Dublin City University and Partners’ Participation in the INS and
  VTT Tracks at TRECVid 2016.
\newblock In: Proceedings of TREVid, NIST, Gaithersburg, Md., USA; 2016.

\bibitem{le2016nii}
Le DD, Phan S, Nguyen VT, Renoust B, Nguyen TA, Hoang VN, et~al.
\newblock NII-HITACHI-UIT at TRECVID 2016.
\newblock In: TRECVID 2016 Workshop. Gaithersburg, MD, USA; 2016.

\bibitem{wahla2016the}
Wahla SQ, Waqar S, Khan MUG, Gotoh Y.
\newblock The University of Sheffield and University of Engineering and
  Technology, Lahore at TRECVid 2016: Video to Text Description Task.
\newblock In: TRECVID 2016 Workshop. Gaithersburg, MD, USA; 2016.

\bibitem{vedantam2015cider}
Vedantam R, Lawrence~Zitnick C, Parikh D.
\newblock {CIDEr}: {C}onsensus-based {I}mage {D}escription {E}valuation.
\newblock In: Proceedings of the IEEE Conference on Computer Vision and Pattern
  Recognition; 2015. p. 4566--4575.

\bibitem{xu2016msr}
Xu J, Mei T, Yao T, Rui Y.
\newblock {MSR-VTT}: {A} {L}arge {V}ideo {D}escription {D}ataset for {B}ridging
  {V}ideo and {L}anguage.
\newblock In: Conference on Computer Vision and Pattern Recognition (CVPR);
  2016.

\bibitem{williams-test}
Williams EJ.
\newblock Regression analysis. vol.~14.
\newblock Wiley New York; 1959.

\bibitem{GrahamBaldwin:14}
Graham Y, Baldwin T.
\newblock Testing for Significance of Increased Correlation with Human
  Judgment.
\newblock In: Proceedings of the 2014 Conference on Empirical Methods in
  Natural Language Processing (EMNLP). Doha, Qatar: Association for
  Computational Linguistics; 2014. p. 172--176.
\newblock Available from: \url{http://www.aclweb.org/anthology/D14-1020}.

\end{thebibliography}

\end{document}